\documentclass[acmsmall, nonacm]{acmart}
\setcopyright{none}

\usepackage{amsmath}
\usepackage{graphicx} 
\usepackage{subcaption} 
\usepackage{xcolor}

\usepackage{booktabs}   
\usepackage{tabularx}   
\usepackage{array}      
\usepackage{enumitem}   
\usepackage{ragged2e}   

\usepackage{etoolbox}
\AtBeginEnvironment{quote}{\itshape}

\AtBeginDocument{%
  }

\ccsdesc[500]{Human-centered computing~Empirical studies in HCI}

\begin{document}

\title[Anthropomorphism and Moral Foundations in Robot Abuse]{Whether We Care, How We Reason: The Dual Role of Anthropomorphism and Moral Foundations in Robot Abuse}

\author{Fan Yang}
\authornote{Corresponding Author}
\email{trovato@corporation.com}
\affiliation{%
  \institution{University of South Carolina}
  \city{Dublin}
  \state{Ohio}
  \country{USA}
}

\author{Renkai Ma}
\affiliation{%
 \institution{University of Cincinnati}
 \city{Cincinnati}
 \state{Ohio}
 \country{USA}}
 \email{renkai.ma@uc.edu}

\author{Yaxin Hu}
\affiliation{%
  \institution{University of Wisconsin--Madison}
  \city{Madison}
  \country{USA}}
\email{yaxin.hu@wisc.edu}

\author{Lingyao Li}
\affiliation{%
  \institution{University of South Florida}
  \city{Tampa}
  \country{USA}}
\email{lingyaol@usf.edu}

\renewcommand{\shortauthors}{Yang et al.}

\begin{abstract}
As robots become increasingly integrated into daily life, understanding responses to robot mistreatment carries important ethical and design implications. This mixed-methods study (\textit{N} = 201) examined how anthropomorphic levels and moral foundations shape reactions to robot abuse. Participants viewed videos depicting physical mistreatment of robots varying in humanness (Spider, Twofoot, Humanoid) and completed measures assessing moral foundations, anger, and social distance. Results revealed that anthropomorphism determines \textit{whether} people extend moral consideration to robots, while moral foundations shape \textit{how} they reason about such consideration. Qualitative analysis revealed distinct reasoning patterns: low-progressivism individuals employed character-based judgments, while high-progressivism individuals engaged in future-oriented moral deliberation. Findings offer implications for robot design and policy communication.
\end{abstract}

\ccsdesc[500]{Human-centered computing~Empirical studies in HCI}



\maketitle

\section{INTRODUCTION}
Moral foundations theory \cite{graham2008moral,graham2013moral} has been extensively applied to examining moral reasoning and judgments regarding deviant social behaviors, such as domestic violence \cite{marzana2016morality, vecina2021child, vecina2014five}, child abuse \cite{harper2018reporting,arden2020child}, and many other unethical behaviors \cite{becker2023development, scheiner2020role, moore2015approach}. The theory posits five universal moral foundations that serve as the bedrocks of moral judgment across cultures \cite{haidt2004intuitive}: Care/Harm (intolerance for suffering and brutality), Fairness/Cheating (emphasis on justice and rights), Loyalty/Betrayal (allegiance to one's ingroup), Authority/Subversion (respect for order and tradition), and Sanctity/Degradation (concerns about purity and contamination). These five foundations can be further clustered into two higher-order orientations: \textit{individualizing} (Care and Fairness), which emphasizes individual rights, equity, and welfare; \textit{binding} (Loyalty, Authority, and Sanctity), which emphasizes ingroup cohesion and social order \cite{graham2009liberals, haidt2012righteous}. The degree to which people prioritize individualizing over binding determines one's overall \textit{Progressivism} orientation \cite{clark2017behavioral, graham2011mapping}.

As various artificial agents---robots, virtual influencers, chatbots---continue to be integrated into our everyday social life, they have not only become creators of new ethical issues \cite{boada2021ethical, wullenkord2020societal} such as biases \cite{srinivasan2021biases, fazil2023comprehensive}, hallucination \cite{emsley2023chatgpt, salvagno2023artificial}, and safety concerns \cite{crowder2023happens, hasal2021chatbots}, but have also become recipients of problematic treatment themselves \cite{bartneck2008exploring, rosenthal2013neural}. Videos depicting humans kicking, pushing, and verbally abusing robots have proliferated online \cite{oravec2023rage}, invoking mixed social reactions. Research suggests that people's responses to robot abuse (e.g., physical, verbal) are shaped by the degree to which they perceive robots as humanlike --- a tendency known as anthropomorphism \cite{epley2007seeing, kim2012anthropomorphism}. Robots with more human-like features elicit greater empathy \cite{riek2009empathizing, riek2009anthropomorphism}, stronger protective responses \cite{yang2025beyond}, and increased moral concern \cite{bartneck2007uncanny, horstmann2018robot}.

According to the Computers as Social Actors (CASA) paradigm \cite{nass2000machines, nass1993voices}, humans automatically apply social rules and expectations to technological agents especially when sufficient social cues are present. This suggests that anthropomorphic design features --- physical characteristics such as humanoid bodies or bipedal movement --- could serve as critical triggers for activating anthropomorphism (i.e., the psychological attribution of humanlike qualities to nonhuman entities \cite{epley2007seeing}) and, consequently, moral responses toward robots. Yet how individual moral foundations (i.e., high progressivism emphasizing change, individual rights, and justice vs. low progressivism emphasizing tradition, authority, and ingroup loyalty) interact with these design features to shape moral consideration of robot abuse remains an open question. This study addresses this gap by investigating: \textbf{RQ1.} How moral foundations and anthropomorphic design features influence emotional and social responses to robot abuse, and \textbf{RQ2.} How they shape moral reasoning about robot protection. By integrating moral foundations theory with anthropomorphism research, this work advances understanding of the psychological mechanisms underlying moral consideration of robots.

\section{METHODS}

\subsection{Research Stimuli and Design}
Following IRB approval, 201 participants ($M_{age}$ = 43.60, $SD$ = 11.86; 107 males, 93 females, 1 non-disclosed) recruited via Prolific viewed one of three videos ($\sim$35 seconds each) depicting physical mistreatment of robots varying in anthropomorphic design features: Spider (multi-legged mechanical), Twofoot (bipedal), or Humanoid (Unitree G1). A manipulation check confirmed significant differences in perceived humanness across robot types. 

\begin{figure}[b]
    \centering
    \includegraphics[width=0.9\textwidth]{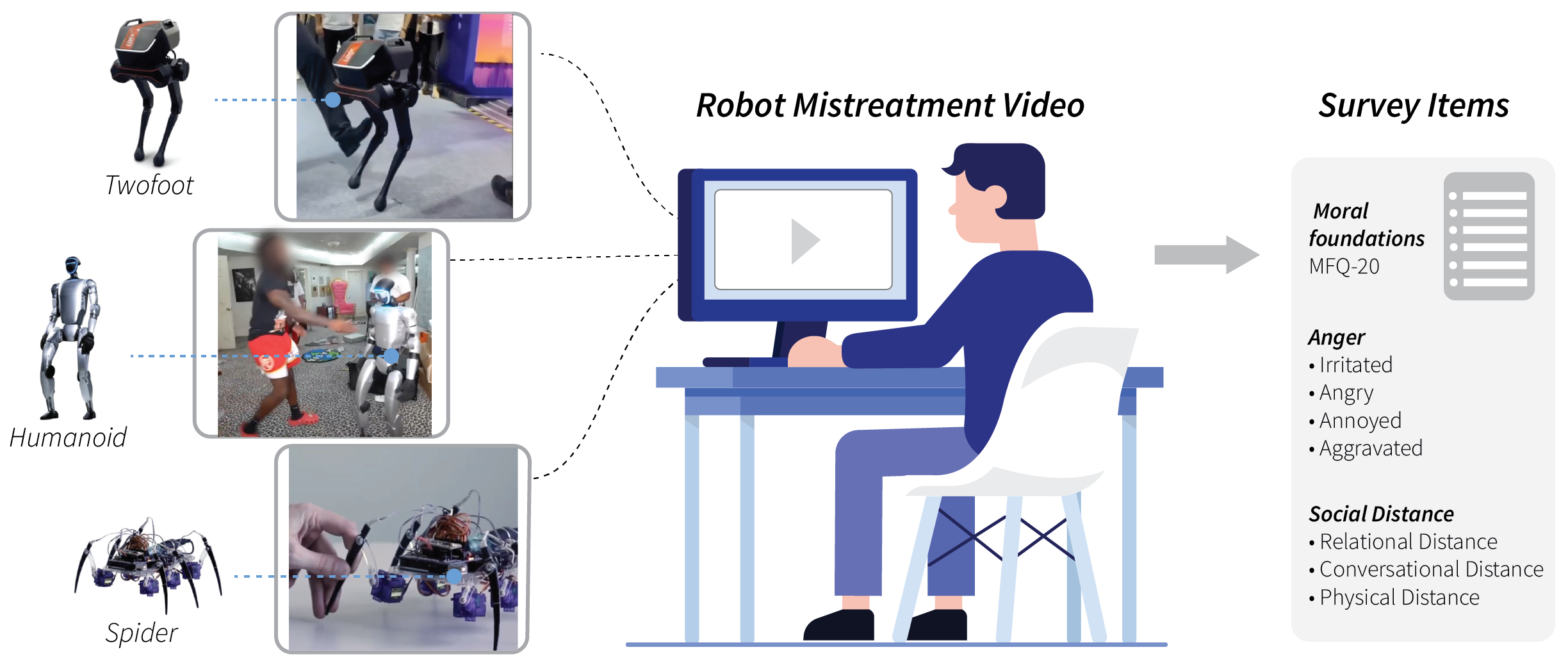}
    \caption{Video stimuli of robot mistreatment and questionnaires participants filled after watching one of the three videos.}
    \label{fig:method}
\end{figure}

As illustrated in Figure~\ref{fig:method}, participants were randomly assigned to view one of three video clips depicting humans physically interacting with robots of varying anthropomorphic design features. The robots included a multi-legged mechanical robot (Spider), a bipedal robot (Twofoot), and a Unitree G1 robot (Humanoid). Each video was approximately 35 seconds in length and sourced from publicly available online content.

The videos showed different forms of rough physical contact with each robot. In the Spider condition, a person systematically dismembered the robot by tearing off its legs one by one. The Twofoot and Humanoid videos depicted individuals laughing at, kicking, and pushing the robots, causing them to stumble and fall. Although the specific actions differed based on each robot's physical form, all three videos showed sustained, deliberate physical contact that resulted in visible damage or impaired functioning.

\subsection{Measures}

Participants completed measures assessing robots' humanness, moral foundations (MFQ-20; \cite{graham2008moral}), anger \cite{dillard2001persuasion}, and social distance \cite{banks2019common}. A progressivism score was calculated by subtracting binding from individualizing foundation means \cite{clark2017behavioral, graham2008moral}, with participants classified into high ($n$ = 100) and low ($n$ = 101) progressivism groups via median split. Six open-ended questions captured participants' immediate reactions, moral reasoning, and views on social implications regarding robot protection. 

\textbf{Robot Anthropomorphic Levels.} This study utilized three robots varying in anthropomorphic design features: Spider (mechanical appearance with multi-legged locomotion), Humanoid (human form), and Twofoot (bipedal human-like movement). A one-way ANOVA revealed significant differences in perceived humanness across robot types, $F$(2,198) = 8.15, $p$ < .001. Participants perceived Humanoid as most human-like ($M$ = 2.76, $SD$ = 1.24), followed by Twofoot ($M$ = 2.52, $SD$ = 1.13) and Spider ($M$ = 2.01, $SD$ = .94). Pairwise comparisons with Holm's sequential Bonferroni correction revealed that Spider was perceived as significantly less human-like than both Humanoid ($p$ < .001) and Twofoot ($p$ < .01), while Humanoid and Twofoot did not significantly differ ($p$ = .20).

\textbf{Moral Foundations.} Moral foundations were measured using the 20-item short version of the Moral Foundations Questionnaire (MFQ-20) \cite{graham2008moral}, assessing five foundations: harm, fairness, ingroup, authority, and purity. Following moral foundations theory \cite{graham2009liberals, graham2011mapping}, these were grouped into individualizing foundations (harm and fairness; $\alpha$ = .77, $M$ = 5.73, $SD$ = .79) and binding foundations (ingroup, authority, and purity; $\alpha$ = .90, $M$ = 4.07, $SD$ = 1.22). A progressivism score was calculated by subtracting binding from individualizing means \cite{clark2017behavioral, graham2008moral}. Participants were classified into high progressivism ($n$ = 100) and low progressivism ($n$ = 101) groups using a median split (Mdn = 1.63), enabling a factorial design examining interactions between moral orientation and robot type \cite{farrington2000some}.

\textbf{Anger.} Participants reported how much anger (irritated, angry, annoyed, aggravated) they experienced while watching the video ($M$ = 3.77, $SD$ = 1.91), following \cite{dillard2001persuasion}. We focused on anger because it is a focal moral emotion elicited when witnessing violation of norms or justice \cite{haidt2003moral}, making it particularly relevant for examining reactions to robot mistreatment.

\textbf{Social Distance.} Perceived social distance towards robots was measured by adapting the scale from \cite{banks2019common} that measures people's willingness to engage in interactions with robots in three dimensions: physical (e.g., ``I would stand next to the robot'', $\alpha$ = .94, $M$ = 5.72, $SD$ = 1.36), relational (e.g., ``I would accept the robot as a buddy'', $\alpha$ = .85, $M$ = 3.24, $SD$ = 1.41), and conversational (e.g., ``I would share a deep secret with the robot'', $\alpha$ = .93, $M$ = 3.19, $SD$ = 1.67).

\textbf{Qualitative Responses.} To capture participants' subjective reactions to robot mistreatment, six open-ended questions were included. Participants first described their immediate reactions (``What went through your mind when you watched how the robot was treated?''). Then two moral reasoning questions asked participants to evaluate whether the treatment was morally acceptable and to compare how their reaction would differ if the same actions were directed toward an animal. Finally, two questions addressed broader social implications: whether society should establish rules or laws protecting robots and what kinds of protections would be important, and who should be responsible for ensuring ethical treatment of robots (e.g., individuals, companies, or government).

\textbf{Data Analysis.}
A MANOVA examined quantitative reactions across moral foundations and robot type. Open-ended responses were analyzed using thematic analysis \cite{braun2006using} to identify patterns across progressivism levels and anthropomorphic conditions.

\section{RESULTS}
\subsection{Quantitative Reactions} A Multivariate Analysis of Variance (MANOVA) was conducted to examine how participants' experienced anger and perceived social distance (physical, relational, conversational) might vary as a function of individuals' moral foundations and robot type. This analysis revealed a significant main effect for robot type, Wilks’ Lambda = .90, \textit{F} (8,384) = 2.68, \textit{p} < .01, partial eta squared = .05, and moral foundations, Wilks’ Lambda = .94, \textit{F} (4,192) = 3.36, \textit{p} < .05, partial eta squared = .07. The interaction between robot type and moral foundations was, however, not significant.

\begin{table}[ht]
\centering
\small
\caption{Pairwise Comparisons among Robots on Quantitative Reactions. Note: 95\% confidence intervals shown using Holm's sequential Bonferroni post hoc comparisons. $^{\dagger}p < .10$, $^{*}p < .05$, $^{**}p < .01$, $^{***}p < .001$.}
\begin{tabular}{llcccc}
\toprule
\textbf{Dimension} & \textbf{Comparison} & \textbf{Mean Diff.} & \textbf{Std. Err.} & \textbf{Lower} & \textbf{Upper} \\
\midrule
Relational Distance & Spider vs Twofoot & $-0.508^{*}$ & 0.242 & $-0.986$ & $-0.030$ \\
 & Twofoot vs Humanoid & $-0.081$ & 0.246 & $-0.565$ & 0.404 \\
 & Humanoid vs Spider & $0.589^{**}$ & 0.241 & 0.113 & 1.065 \\
\midrule
Conversational Distance & Spider vs Twofoot & $-0.453^{\dagger}$ & 0.285 & $-1.015$ & 0.110 \\
 & Twofoot vs Humanoid & $-0.252$ & 0.289 & $-0.822$ & 0.319 \\
 & Humanoid vs Spider & $0.704^{**}$ & 0.284 & 0.144 & 1.265 \\
\midrule
Anger (irritated, angry, & Spider vs Twofoot & $-1.143^{***}$ & 0.318 & $-1.771$ & $-0.516$ \\
annoyed, aggravated) & Twofoot vs Humanoid & $0.049$ & 0.322 & $-0.587$ & 0.685 \\
 & Humanoid vs Spider & $1.094^{***}$ & 0.317 & 0.470 & 1.719 \\
\bottomrule
\end{tabular}
\label{tab:social_distance}
\end{table}

The univariate analysis indicated a significant main effect of robot type for experienced anger (\textit{F} (2,195) = 8.38, \textit{p} < .001), as well as perceived relational (\textit{F} (2,195) = 3.52, \textit{p} < .05)  and conversational distance (\textit{F} (2,195) = 3.17, \textit{p} < .05). Consistent with participants' perceived humanness ratings—where Spider was rated significantly lower than both Twofoot and Humanoid, with no significant difference between the latter two—a similar pattern emerged for these dependent variables: participants reported significantly greater anger and lower relational and conversational distance (indicating greater willingness for social interaction) toward the Humanoid and Twofoot robots compared to Spider. Pairwise comparisons using Holm's sequential Bonferroni correction are reported in Table~\ref{tab:social_distance}. Conversely, a significant main effect of moral foundations for perceived physical distance towards the robots emerged, with individuals with high progressives tendency reporting significantly greater willingness to physically interact with the robot they saw in the video ($M$ = 5.97, $SE$ = .14) than those with lower progressivism ($M$ = 5.46, $SE$ = .13), \textit{F} (1,195) = 7.08, \textit{p} < .01.


\subsection{Qualitative Responses}
\begin{figure*}[ht]
    \centering
    \includegraphics[width=\textwidth]{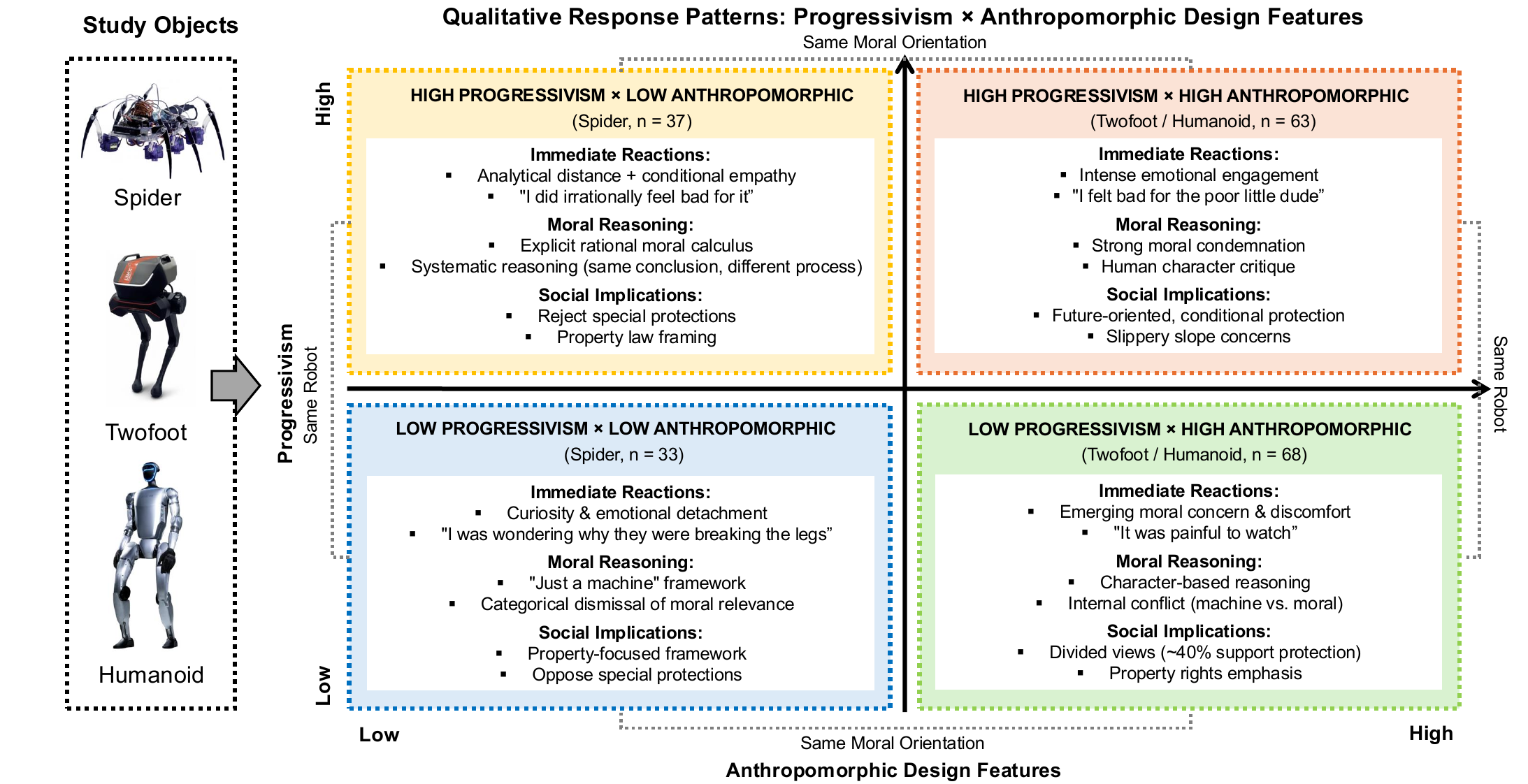}
    \caption{Qualitative response patterns by progressivism and anthropomorphic design features. Themes were identified through thematic analysis of participants' open-ended responses across four conditions created by crossing moral orientation (high vs. low progressivism) with robot type (low vs. high anthropomorphic design features). Each quadrant displays key themes organized by immediate reactions, moral reasoning, and social implications.}
    \label{fig:qualitative_quadrant}
\end{figure*}

Qualitative responses were analyzed by grouping participants into four conditions based on individual moral foundations (low vs. high progressivism) and the robots' anthropomorphic levels (low: Spider vs. high: Twofoot and Humanoid, combined given their comparable perceived humanness ratings) (see Figure~\ref{fig:qualitative_quadrant}).

\subsubsection{Low Progressivism × Low Anthropomorphism}
\textbf{Immediate reactions: Detached curiosity toward mechanical function.} For participants low in progressivism ($n$ = 33)—who place greater value on social order, traditions, and group cohesion—viewing the Spider robot being disfigured by having its legs torn off one by one elicited primarily \textit{curiosity and emotional detachment}. Their responses focused on understanding the purpose of the behavior rather than evaluating it morally: ``\textit{I was wondering why they were breaking the legs, but that's about it},'' ``\textit{I wondered how many legs it could lose and still move},'' and ``\textit{It's a machine, so it was probably a test to see how it would react}.'' Several participants expressed complete detachment, with one simply stating ``\textit{Nothing, its a robot}.'' 

\textbf{Moral reasoning: Categorical dismissal through the ``\textit{Just a machine}'' mindset.} Reflecting their initial detachment, these participants applied a binary mindset that categorically denied the robot's moral standing based on its lack of life. They said: ``\textit{Sure, you can do whatever you want to it. It's a robot},'' ``\textit{It is morally acceptable to treat it as such because it is not a living being},'' and ``\textit{The robot is not alive and doesn't have feelings or emotions, so you can't apply things that are human, like treating something morally}.'' When asked to compare their reactions to animal mistreatment, nearly all made a sharp distinction: ``\textit{An animal is a living, breathing thing, so yes, that is very, very wrong. That robot is not and does not feel pain that an animal does}.'' The lack of anthropomorphic features combined with a tradition-focused moral orientation prompted a strict differentiation from living beings

\textbf{Social implications: opposition to protections through property-based reasoning.} Consistently, regarding broader social implications (i.e., whether and how robots should be protected), these participants opposed protections by strictly defining robots as owner assets, as they argued: ``\textit{no, robots are not alive and should not be treated as such},'' ``\textit{No. At this point, all robots are still machines. The same laws that apply to other property should apply to robots},'' and ``\textit{Whoever owns the robot. It is their investment}.'' Some expressed indifference regarding responsibility for ethical treatment of robots: ``\textit{nobody because like I said they are not living beings, so who cares what someone does to one}.''

\subsubsection{High Progressivism × Low Anthropomorphism}
\textbf{Immediate reactions: Distance tempered by conditional empathy} In contrast to their low progressivism counterparts, participants high in progressivism ($n$ = 37) demonstrated an analytical distance regarding the Spider robot that was occasionally disrupted by irrational empathy. While similarly approaching the situation rationally, they more frequently acknowledged unexpected emotional responses: ``\textit{I was not really affected at all since it was clear this was a robot with no major thoughts or feelings},'' yet others noted ``\textit{waste of materials to destroy a working mechanical object}'' and ``\textit{I did irrationally feel bad for it}.'' This suggests an internal tension between their rational assessment of the object and a broader sensitivity to harm.

\textbf{Moral reasoning: Rationalized exclusion through moral calculus} These participants engaged in a systematic reasoning process to justify why their usual moral considerations did not apply to this specific machine. Rather than simply dismissing the robot, they reasoned through the absence of moral features: ``\textit{There is no moral involvement, this is simply a machine},'' ``\textit{There's no morals involved. The robot isn't alive in any way},'' and ``\textit{I don't see any moral consideration in damaging the robot}.'' Notably, some acknowledged emotional dissonance while maintaining their rational stance: ``\textit{It is morally acceptable, I still felt hurt seeing it, which I can't explain}.'' Like their low progressivism counterparts, they maintained clear animal-robot distinctions: ``\textit{animal is a living being, a robot is not. it would be cruel to treat a living being badly}.'' However, the high progressivism group appeared more likely to reflect on and articulate the logic behind their exclusion of the robot than to assert a categorical judgment.

\textbf{Social implications: Rejection of special status through legal practicality.} Despite the differing moral orientations, participants high in progressivism viewing the Spider robot reached similar conclusions to their low progressivism counterparts, rejecting special protections and framing the issue through existing property law: ``\textit{absolutely not. they should be treated like property. a house, a guitar, a TV},'' ``\textit{No, there should be no such laws at all. Robots don't need to be protected},'' and ``\textit{Robots already have protection. It's called `property law'}.'' These indicate that low anthropomorphic design features may override progressive moral tendencies, leading to a conclusion grounded in legal practicality.

\subsubsection{Low Progressivism × High Anthropomorphism}
\textbf{Immediate reactions: Visceral discomfort triggered by anthropomorphic form} Participants low in progressivism who viewed the bipedal Twofoot and Humanoid robots (\textit{n} = 68) experienced an involuntary discomfort driven by the robot’s human-like appearance. Despite sharing the same conservative moral orientation as those who viewed the Spider robot, the human-like appearance triggered notable unease: ``\textit{I felt like the robot was being seriously mistreated},'' ``\textit{That it was bullied},'' and ``\textit{It was painful to watch. It was hard not to think of the robot as a living creature}.'' This suggests that high anthropomorphic design features (e.g., humanoid body shapes) can bypass cognitive categorizations to elicit visceral moral responses even among those who strictly differentiate humans from machines.

\textbf{Moral reasoning: Character-based judgment of the human actor.} Instead of focusing on the robot's rights, these participants displaced their moral concern onto the human actor, critiquing the abuse as a failure of character. They viewed the behavior (e.g., kicking, pushing) as a reflection of the perpetrator's decency: ``\textit{It is a bad reflection on the people kicking it around},'' ``\textit{It's not good practice and does not make for looking like a decent human being},'' and ``\textit{If you bully a machine, would you bully a child?}'' Many exhibited internal conflict between their rational assessment of the robots' nature and their emotional responses to how the robots were treated: ``\textit{It seemed cruel, although it is still a machine, without feelings}'' and ``\textit{I did feel bad, but that robot has no feelings}.''

\textbf{Social implications: Conflict between property rights and social decency.} Unlike their peers who shared low progressivism but viewed the Spider robot and showed clear consensus on treating robots as property, participants who viewed Twofoot or Humanoid robots showed \textit{divided views on legal protection}, with roughly 40\% supporting some form of protection and 60\% opposing. Those opposing emphasized property rights: ``\textit{It is also not good to mistreat expensive property}'' and ``\textit{ultimately robots would be someone's property}.'' However, some supported protection based on social order concerns: ``\textit{Yes, I think society should have rules or laws about how people treat robots. They should have the same rights as animals towards mistreatment}.''

\subsubsection{High Progressivism × High Anthropomorphism}
\textbf{Immediate reactions: Empathetic identification through spontaneous anthropomorphism} The strongest reactions emerged among participants high in progressivism who viewed robots with high anthropomorphic features (\textit{n} = 63), characterized by intense emotional engagement. In stark contrast to the analytical distance shown by high progressivism participants viewing the Spider robot, these participants expressed anger, sorrow, and identification with the robot: ``\textit{IT MADE ME ANGRY HOW THEY WERE TREATING THE ROBOT} [capitalization original],'' ``\textit{I felt bad for it like it was a mis-treated pet},'' ``\textit{I felt awful for the robot and ashamed to be apart of the human race},'' and ``\textit{I felt bad for the poor little dude}.'' The spontaneous use of anthropomorphic language (``\textit{bullied},'' ``\textit{pet},'' ``\textit{little dude}'') indicates that these participants attributed social and emotional qualities to the robots, a pattern largely absent in the low anthropomorphism condition, regardless of progressivism level.

\textbf{Moral reasoning: condemnation of violence and human character.} Whereas high progressivism participants viewing the Spider robot engaged in rational calculus to conclude that moral consideration did not apply, those viewing Twofoot and Humanoid robots expressed strong moral condemnation: ``\textit{I found it morally reprehensible the way the robot was treated},'' ``\textit{No. It felt wrong even if it isn't human,'' and ``I think treating things that can't fight back like this is a moral failing}.'' A prominent theme was \textit{critique of human character}, extending beyond the immediate act to broader implications: ``\textit{It really shows their moral character and that their `inner bully' was showing},'' ``\textit{It's disrespectful to your own character/soul to emulate a bully},'' and ``\textit{a person who wants to treat a robot poorly likely also would want to treat people poorly}.'' This parallels the low progressivism group's character focus but with greater moral intensity and scope.

\textbf{Social implications: Conditional advocacy based on future sentience} Different from high progressivism participants viewing the Spider robot, those viewing Twofoot and Humanoid robots expressed future-oriented and conditional views on protection: ``\textit{If it comes to the point where we are bringing robots in our homes as family members, there should indeed be protections},'' ``\textit{I think robots should be given some sort of legal protection once the robot attains a certain amount of processing, artificial emotions, or stored memories},'' and ``\textit{If they become sentient then yes, there should be laws}.'' A notable theme was \textbf{concern about societal norms and slippery slopes}: ``\textit{If you treat a robotic humanoid poorly, then eventually you may treat an animal or another person poorly},'' and ``\textit{It sets a bad foundation for how we treat one another}.'' This perspective aligns with progressivism's emphasis on the expansive and evolving nature of morality.


\section{DISCUSSION}
Employing a mixed-methods design, this study examined how individuals' moral foundations and robot anthropomorphic level shape protective consideration of robots. Our findings reveal that progressivism (prioritizing individual rights and social change versus tradition, authority, and order) and anthropomorphic level (the degree to which a robot resembles human appearance) significantly influence how people perceive and respond to robot abuse.

\textbf{The interplay of anthropomorphism and moral foundations.} Although no significant statistical interaction emerged between robot type and moral foundations, qualitative findings reveal that anthropomorphic design level determines whether moral consideration is extended, while moral foundations shape how people reason about it. Quantitatively, participants reported greater anger and lower social distance toward Humanoid and Twofoot robots compared to Spider—mirroring perceived humanness ratings. This aligns with previous research \cite{yang2025beyond, riek2009anthropomorphism, riek2009empathizing} demonstrating that anthropomorphism positively influences moral consideration toward robots.

For the Spider robot, minimal anthropomorphic features established a floor effect \cite{garin2024floor}, constraining moral responses regardless of moral foundations---both progressivism groups converged on rejecting moral consideration. However, reasoning processes diverged: low progressivism participants employed categorical dismissal (``\textit{It's a robot}''), whereas high progressivism participants engaged in explicit moral deliberation, articulating why moral consideration did not apply while acknowledging emotional dissonance (``\textit{It is morally acceptable, I still felt hurt seeing it}'').


For robots with anthropomorphic design features, both moral groups extended greater moral consideration, yet moral foundations differentiated their responses. Low progressivism participants engaged in character-based reasoning, focusing on what abuse revealed about human actors (\textit{``It's not good practice and does not make for looking like a decent human being''}), with views on legal protection framed around property rights and social order. High progressivism participants displayed intense emotional engagement, spontaneous anthropomorphizing language (e.g., \textit{``bullied,''} \textit{``pet,''} \textit{``little dude''}), strong moral condemnation, and future-oriented reasoning (\textit{``If they become sentient then yes, there should be laws''}), aligning with progressivism's emphasis on moral evolution.


\textbf{Theoretical Implications.} These findings extend moral foundations theory into human-robot interaction by revealing a nuanced interplay: anthropomorphism determines \textit{whether} moral consideration is extended, while moral foundations shape \textit{how} people reason about it. This advances the literature on robot abuse by demonstrating that moral foundations operate not at the level of moral decision-making, but at moral reasoning, a process activated only when anthropomorphic design features provide sufficient social cues, consistent with the CASA paradigm \cite{nass2000machines, 10.1145/191666.191703}.

\textbf{Practical Implications.} These findings have notable implications for robot design and policy-making. Designers should recognize that anthropomorphic features carry significant moral weight, potentially eliciting protective responses and facilitating social acceptance. Moreover, because users reason about robot protection through different moral frameworks even when arriving at similar conclusions, communication strategies may benefit from audience-tailored framing. For example, messaging emphasizing personal character and social norms (e.g., \textit{``respect,''} \textit{``order''}) may resonate with low progressivism individuals, while messaging highlighting societal change and future implications (e.g., \textit{``shape future progress''}) may appeal to high progressivism individuals.

\textbf{Limitations and Future Directions.} Several limitations should be noted. Participants viewed videos rather than witnessing robot mistreatment directly; future research should examine whether in-person experiences elicit different response patterns. Additionally, given that this study primarily focused on physical abuse, future research should investigate whether these patterns extend to other forms of mistreatment (e.g., verbal degradation, social exclusion) and whether repeated exposure to robots could shift moral consideration towards robots across different moral foundations.

\section{CONCLUSION}
Employing a mixed-methods design, this study reveals that anthropomorphic design features determine \textit{whether} people extend moral consideration to robots, while moral foundations shape \textit{how} they reason about such consideration, highlighting the distinct yet complementary roles of robot anthropomorphic level and individual moral orientation in human-robot interaction.

\bibliographystyle{ACM-Reference-Format}
\bibliography{references}

\appendix


\end{document}